\documentclass{article}
\usepackage{amsmath,amssymb,amsfonts}
\usepackage{algorithmic}
\usepackage{graphicx}
\usepackage{textcomp}
\usepackage{tabularx}
\usepackage{xcolor}
\usepackage{amsthm}
\newtheorem{definition}{Definition}
\usepackage{enumitem}
\usepackage{lipsum}
\usepackage{indentfirst}
\usepackage{microtype}
\usepackage{geometry}
\usepackage{natbib} 
\title{Turbulence Regression}

\author{
Yingang Fan \\
College of Computer and\\
Information Science,\\
Southwest University,\\
Chongqing, China,\\
fanyingang2003@163.com
\and
Binjie Ding* \\
College of Computer and\\
Information Science,\\
Southwest University,\\
Chongqing, China,\\
yoooou.ding@gmail.com
\and
Baiyi Chen \\
College of Information Security, \\
Chongqing College of\\
Mobile Communication,\\
Chongqing, China,\\
chenbaiyiaz97@163.com
}

\date{}

\begin{document}

\maketitle

\begin{abstract}
Air turbulence refers to the disordered and irregular motion state generated by drastic changes in velocity, pressure, or direction during airflow. Various complex factors lead to intricate low-altitude turbulence outcomes. Under current observational conditions, especially when using only wind profile radar data, traditional methods struggle to accurately predict turbulence states. Therefore, this paper introduces a NeuTucker decomposition model utilizing discretized data. Designed for continuous yet sparse three-dimensional wind field data, it constructs a low-rank Tucker decomposition model based on a Tucker neural network to capture the latent interactions within the three-dimensional wind field data. Therefore, two core ideas are proposed here: 1) Discretizing continuous input data to adapt to models like NeuTucF that require discrete data inputs. 2) Constructing a four-dimensional Tucker interaction tensor to represent all possible spatio-temporal interactions among different elevations and three-dimensional wind speeds. In estimating missing observations in real datasets, this discretized NeuTucF model demonstrates superior performance compared to various common regression models.
\end{abstract}


\section{Introduction}
Atmospheric turbulence refers to an unordered and irregular state of motion generated by drastic changes in velocity, pressure, or direction during airflow. As a critical and highly complex phenomenon in atmospheric science, its accurate prediction holds profound significance in fields such as meteorology and aviation \cite{AviationTurbulence, BarnesTurbulence}.

However, the nonlinearity of atmospheric turbulence poses severe challenges to traditional analysis and modeling methods\cite{StrakaNonlinear, RenANNSubgrid}. Under existing observation conditions, when relying solely on wind profile radar data, traditional methods struggle to capture the potential nonlinear interactions in continuous and sparse three-dimensional wind field data\cite{ChengLidarTurbulence}, hindering the application of atmospheric turbulence in meteorology and aviation\cite{TenenbaumJetTurbulence} .

To address these challenges, this paper proposes a NeuTucker model based on discretized data, which constructs a low-rank Tucker decomposition model using a Tucker neural network to capture the latent interactions within the three-dimensional wind field data. The Richardson number $(Ri)$ is a core metric for diagnosing the occurrence of turbulence and stability. This study leverages $Ri$ to establish a bridge between theoretical turbulence diagnosis and data-driven modeling. The main contributions of this paper are as follows:

1. Discretizing continuous input data to adapt to models that require discrete data inputs, such as NeuTucker.

2. Constructing a four-dimensional Tucker interaction tensor that can encode all potential spatiotemporal interactions between different altitudes and the three wind components $(h,u, v, w)$.

In practical applications, this discretized NeuTucker model outperforms common regression models in estimating missing observations in real datasets. This research advances the integration of discrete data processing and tensor decomposition in atmospheric turbulence, providing a feasible solution for improving turbulence prediction and holding direct significance for fields such as meteorology and aviation.

\section{Preliminaries}

\subsection{Richardson Number}
The Richardson Number ($Ri$) is a core indicator for diagnosing the occurrence and development trend of turbulence\cite{ShaoRichardson} . It is often used to quantify the relative importance of the atmospheric buoyancy effect and the wind shear effect, thereby judging the stability of turbulence . Essentially, it is the ratio of buoyancy work to shear work, and its value directly reflects the possibility of turbulence occurrence, making it one of the key indicators for measuring turbulence intensity in the field of turbulence research .

The calculation formula of the Richardson Number\cite{Freire2019} is:
\begin{equation}
    Ri = \frac{g}{\theta} \frac{\partial \theta}{\partial z} /\left(\left|\frac{\partial u}{\partial z}\right|^{2}+\left|\frac{\partial v}{\partial z}\right|^{2}\right)
    \label{eq:Ri}
\end{equation}

It is generally believed that when $(Ri < 0.25)$, turbulence is prone to occur and is relatively strong, and when $(Ri > 1.0)$, turbulence is relatively stable\cite{Freire2019} . This critical range provides an important basis for judging low-altitude turbulence monitoring.

\subsection{Discretization}
In atmospheric turbulence modeling, the three-dimensional wind components and elevation$(u,v,w,h)$ data are key features, whose continuous nature critically affects model accuracy \cite{r27,r28,r29}. Discretization divides the continuous feature space into finite intervals, allowing the model to learn specific weights for different ranges and better capture nonlinear patterns \cite{r30,r31}.

Furthermore, continuous wind field data are prone to measurement noise and outliers \cite{r32,r33,Chakravarty2022}. Discretization bins extreme values into boundary intervals, reducing their adverse effects on training and improving feature stability. This approach particularly enhances the model's identification of critical transition regions under low wind speed or strong shear conditions \cite{r34,r35}.

Additionally, discretized features can be processed as categorical variables via embedding layers \cite{r36,r37}. Compared to direct continuous value processing, embedded representations capture latent semantic relationships between features, providing richer information for deep learning models \cite{r38,r39}.

\subsection{Tensor and Discretized Tensor}
Tensors can represent various kinds of data, especially those with complex spatiotemporal interactions among entities \cite{Tang2025Auto,r40,r41,r42,r10,r11,r18,r19,r20}. There, a discretized vector and a mode-4 tensor are defined as follows.

\begin{figure}
    \centering
    \includegraphics[width=8cm]{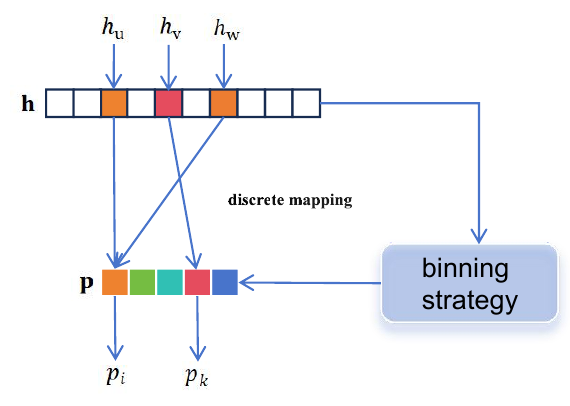}
	\caption{Process of continuous vector discretization, where features within the same value-segmented interval are mapped to an identical integer label}
	\label{fig_Discretization}
\end{figure}

\begin{definition}[\textbf{Discretized Vector}]
	Given a continuous feature vector $\mathbf{h} \in \mathbb{R}^H$, the number of bins is first preset as a bins index vector  $\mathbf{p} \in \mathbb{R}^P$ according to Equation \ref{B_k}. Using a quantile-based strategy as Equation \ref{index}, the continuous values of each feature $h_i$ are divided into intervals $p_k$ containing approximately equal numbers of samples. Each interval is then mapped to an integer label$(1,2,3,\dots, P)$ ,resulting a discretized vector:
    \begin{equation}
         \boldsymbol{\mathcal{D}}(\mathbf{h} )= \mathbf{p}
         \label{Discretization}
         \end{equation}
   
\end{definition}
Where $ \boldsymbol{\mathcal{D}}(\cdot)$ denotes the discretization function that maps continuous values into discrete ones, as illustrated in Fig \ref{fig_Discretization}. 

\begin{definition}[\textbf{Mode-4 Tensor}]
	Given four positive integer sets $H = \{h: h \in \mathbb{R}\}$, $U = \{u:u \in \mathbb{R}\}$ and $V= \{v: v \in \mathbb{R}\}$, $W= \{w: w \in \mathbb{R}\}$ ,$\boldsymbol{\mathcal{Y}}^{|H| \times |U| \times |V| \times|W|}$ is a mode-4 tensor of which each element $y_{huvw}  \in \mathbb{R}$ denotes an observation of the spatiotemporal interaction of the involved height and the three-dimensional wind field $h, u, v$, and $w$ \cite{r43,r44,r45}.
\end{definition}

In turbulence prediction, wind data are often extremely incomplete. Thus, three-dimensional wind field data tensor can be represented by a High Dimensional and Incomplete(HDI) tensor with three spatial modes and one height mode \cite{r46,r47,r25,r26}.

\begin{definition}[\textbf{Discretized Tensor}]
	Given a mode-4 continuous tensor $\boldsymbol{\mathcal{Y}}^{|H| \times |U| \times |V| \times|W|}$, we apply the discretization process defined in Definition $1$ to the continuous feature vectors of each of its modes. Preset the number of bins for each mode, generating the corresponding bin index vectors $ \mathbf{p} \in \mathbb{R}^P,\mathbf{i}\in\mathbb{R}^I, \mathbf{j}\in\mathbb{R}^J, \mathbf{k}\in\mathbb{R}^K$. Then, we apply the Equation \ref{Discretization} to obtain :
    \begin{equation}
      \boldsymbol{\mathcal{Z}}= D(\boldsymbol{\mathcal{Y}})
    \end{equation}
    
\end{definition}

Where $\boldsymbol{\mathcal{Z}} \in\mathbb{R}^{|P| \times |I| \times |J| \times|K|}$ and each element $\boldsymbol{\mathcal{Z}}_{huvw}$ is no longer a continuous real value but a tuple composed of four discrete integer labels. This discretized tensor $\boldsymbol{\mathcal{Z}}$ preserves the four-dimensional structure of the original continuous tensor $\boldsymbol{\mathcal{Y}}$, which is illustrated in fig \ref{fig_model}, but transforms the continuous physical quantity observations into discrete category labels, thereby preparing the data for subsequent four-dimensional tensor decomposition algorithms \cite{r48,r49,r12,r13,r14,r21,r22}.

\section{Turbulence Regression}

\subsection{Data Processing}\label{AA}
We adopt an equal-frequency binning strategy to discretize the wind speed components and elevation $(u, v, w, h)$ into $k$ intervals, ensuring a balanced sample distribution across the discretized category indices into low-dimensional dense vectors \cite{r50,r51}.

First, the original wind field and elevation data are standardized using the following process:
\begin{equation}
X_{std} = \frac{X - \mu}{\sigma}\label{X_std}
\end{equation}
 where $\mu$ is the feature mean and $\sigma $ is the standard deviation. Standardization ensures that all features are on a comparable scale,  laying the groundwork for subsequent binning \cite{r52,r53}.

 Next, the quantile-based binning method is applied to the standardized continuous features. This ensures that each bin contains approximately the same number of samples \cite{r54,r55}.
 Given the number of bins $k$, the bin boundaries are determined as follows:
 \begin{equation}
     B_k = Q\left(\frac{k}{K}\right), \quad k = 0,1,\ldots,K\label{B_k}
 \end{equation}
in which  $Q(p)$ represents the p-th quantile of the feature values. This method adapts to potential skewness in the data distribution and prevents some intervals from having too few samples \cite{r56,r57}.

 Finally, the continuous feature values are mapped to discrete indices using the following piecewise function, as demonstrated in Fig \ref{Discretization}:
 \begin{equation}
     \text{index} = 
\begin{cases} 
0 & \text{if } x < B_1 \\
i & \text{if } B_i \leq x < B_{i+1} \\
K & \text{if } x \geq B_K
\end{cases}\label{index}
 \end{equation}
where $x$ represents the standardized continuous feature value, $b_i$ denotes the i-th bin boundary, $K$ is the total number of bins, and $index$ is the resulting discrete index value, ranging from $0$ to $k$ \cite{r58,r59}.

\subsection{Neural Tucker Factorization}
The Tucker decomposition is a form of higher-order principal component analysis \cite{r1,r2,r3,r4,r5,r60,r61}. It decomposes an N-th order tensor into a core tensor multiplied by a set of factor matrices along each mode. For a third-order tensor, the Tucker decomposition can be expressed as:
\begin{equation}
    \mathcal{Y} \approx \boldsymbol{\mathcal{G}} \times_1 \mathbf{A} \times_2 \mathbf{B} \times_3 \mathbf{C} = [\![\boldsymbol{\mathcal{G}}; \mathbf{A}, \mathbf{B}, \mathbf{C}]\!]
\end{equation}
    where ${\boldsymbol{\mathcal{G}}}$ is the core tensor, whose elements encode the interactions between different factors, and  $\mathbf{A} \in \mathbb{R}^{I \times P}$, $\mathbf{B} \in \mathbb{R}^{J \times Q}$ and $\mathbf{C} \in \mathbb{R}^{K \times R}$ are the factor matrices, which can be regarded as latent dimensions along each mode \cite{r62,r63,r23,r24}. 
    
    The element-wise calculation formula is given by:
    \begin{equation}
        \hat{y}_{ijk} = \sum_{p=1}^{P} \sum_{q=1}^{Q} \sum_{r=1}^{R} g_{pqr} a_{ip} b_{jq} c_{kr}
    \end{equation}
    
    this indicates that each element of the final tensor is a weighted sum of all possible combinations of latent features, with the weights provided by the core tensor  ${\boldsymbol{\mathcal{G}}}$ .
\begin{figure}
    \centering
    \includegraphics[width=8cm]{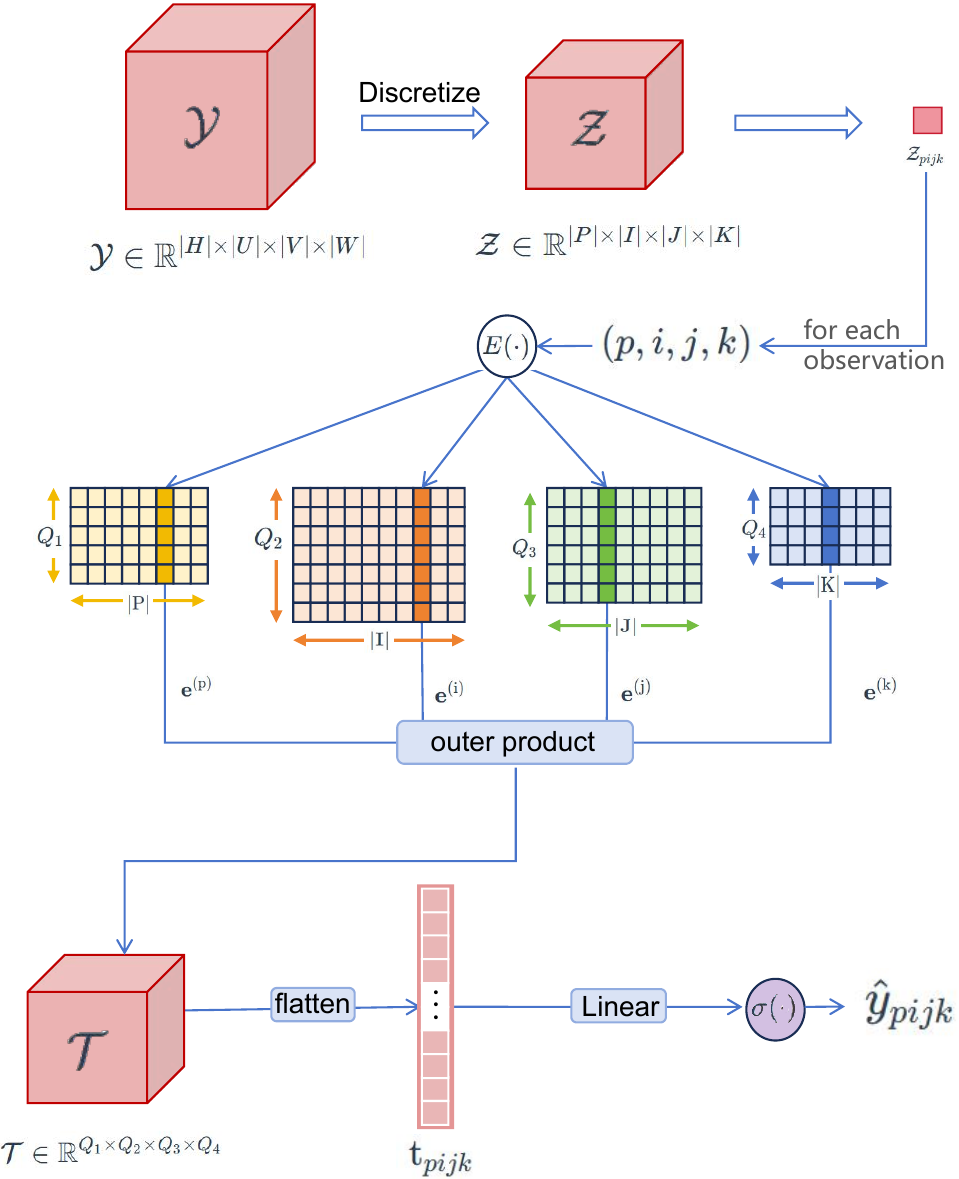}
	\caption{An illustration of the discrete Tucker decomposition process of a four-dimensional continuous tensor.}
	\label{fig_model}
\end{figure}    
\subsection{Regression Prediction}
Inspired by NeuTucF, we propose a Tucker decomposition model based on discretized wind field data.\cite{r86,r87,r88,r89,r90} Under the data conditions of the three-dimensional wind field, we can transform continuous vectors$(\mathbf{h},\mathbf{u},\mathbf{v},\mathbf{w})$ into discretized vectors$(\mathbf{p},\mathbf{i},\mathbf{j},\mathbf{k})$ using Equation \ref{Discretization}, thus preparing for subsequent four-dimensional tensor $ \boldsymbol{\mathcal{Z}}$ decomposition.

Then, for each observation of $ \boldsymbol{\mathcal{Z}}$ illustrated in Fig \ref{fig_model} the discretized four-dimensional input  is transformed into dense vector representations via embedding layers:
\begin{equation}
	\mathbf{e}^{(\mathrm{p})}= E(\mathbf{p}_i)
	\label{eq:embed}
\end{equation}
where $E(\cdot)$ denotes the embedding function that maps discrete values $\mathbf{p}_i$ into latent factor vectors $\mathbf{e}^{(\mathrm{p})}\in\mathbb{R}^{Q_1}$ for the $q$-th dimension.
     
 In the Tucker interaction layer, interaction tensors are constructed via outer product operations:
 \begin{equation}
     \mathcal{T}_{pijk} = \mathbf{e}^{(\mathrm{p})} \circ \mathbf{e}^{(\mathrm{i})}  \circ \mathbf{e}^{(\mathrm{j})}  \circ \mathbf{e}^{(\mathrm{k})}
 \end{equation}
where $\circ$ represents the outer product operator. The resulting four-order interaction tensor $\mathcal{T}_{pijk} \in \mathbb{R}^{Q_1\times Q_2\times Q_3\times Q_4}$ captures all possible feature combinations.

  The interaction tensor is flattened and then passed through a linear layer to realize the weighting effect of the core tensor:
  \begin{equation}
      \hat{y}_{pijk} = \sigma\left(\mathbf{W}^T \cdot \mathcal{F}(\mathcal{T}_{pijk})\right)
  \end{equation}
 where $\mathcal{F}(\cdot)$ is the flatten operation. The weight of the linear layer parameter $\mathbf{W}^T$ is functionally equivalent to the flattened core tensor ${\boldsymbol{\mathcal{G}}}$, and the sigmoid $\sigma(\cdot)$ activation function ensures that the output range remains dimensionally consistent with the normalized data \cite{r64,r65,r66,r67,r68,r69,r75,r76,r77,r78,r79,r80,r81,r82,r83,r84,r85}.

\section{Empirical Studies}
\subsection{General Setting}\label{SCM}
All  experiments were performed on a platform with a 2.50-GHz Intel(R) Core(TM) i5-10300H CPU and one NVIDIA GeForce RTX 2060 GPU with 16-GB RAM. The Richardson number data in the three-dimensional wind field dataset is processed from wind profile radar products and microwave radar data, calculated according to Equations \eqref{eq:Ri}. The wind components and elevation data $(h,u,v,w)$ are obtained from wind profile radar observations, where the wind components are processed using the least squares method. Finally, the two datasets are merged via interpolation along time, height, and station dimensions, resulting in a three-dimensional wind field dataset containing $(h,u,v,w)$ and the Richardson number. The known count of the dataset is 373, and the dimension is $41\times361\times360\times89$. Due to the highly skewed distribution and large variance of the Richardson number, which contradicts the probabilistic assumptions of low-rank decomposition, we apply logarithmic transformation and standardization to the data. This processing aims to make the obtained data more closely resemble a normal distribution, thereby better aligning with the assumptions.

The dataset is partitioned using five fold cross-validation.\cite{r70,r71,r72,r73,r74} The results from the five validation runs are averaged to mitigate the random error introduced by a single data partition. For model evaluation, a classic suite of regression metrics is adopted: Mean Absolute Error $(MAE)$, Root Mean Square Error $(RMSE)$, and Coefficient of Determination $(\mathrm{R}^2)$. \cite{r15,r16,r17}Through the synergistic evaluation of these multi-dimensional metrics, the predictive performance and generalization capability of the model are objectively reflected.\cite{r6,r7,r8,r9}

\begin{table}
\centering
\caption{The summary of results}
\label{tab:results_summary}
\begin{tabular}{cccc}
\hline
\textbf{Model} & \textbf{MAE}              & \textbf{RMSE}             & \textbf{$R^2$}\\ \hline
\textbf{M1}    & \textbf{11.6400 ± 2.4723} & \textbf{20.1061 ± 3.7569} & \textbf{0.3530 ± 0.1953}      \\
M2             & 12.3250 ± 1.9904          & 21.0975 ± 2.9664          & 0.0656 ± 0.1517               \\
M3             & 11.8235 ± 1.1474          & 20.7110 ± 2.6165          & 0.1767 ± 0.0800               \\
M4             & 13.2511 ± 1.3060          & 22.6092 ± 2.0347          & 0.0807 ± 0.1182               \\
M5             & 12.6271 ± 2.2220          & 21.3449 ± 4.7098          & 0.0113 ± 0.1031               \\ \hline
\end{tabular}
\end{table}

\subsection{Performance Evaluation}
Our model is compared with several  models that  commonly perform basic predictions on turbulence analysis. The compared models include: 1) $M1$: The NeuTucF model, based on Tucker decomposition, which captures high-order feature interactions through tensor outer products;
2) $M2$: The MLP (Multi-Layer Perceptron) model, which learns implicit feature relationships via non-linear transformations of concatenated embeddings;
3) $M3$: The Linear, a pure linear model that performs only a weighted summation of feature embeddings without considering interactions;
4) $M4$: The Pairwise focuses on pairwise feature interactions, which processes interaction results through linear layers and concatenates them with original features;
5)$ M5$: The FFN fusing explicit interactions (both pairwise and triple-wise) with deep network features, with output range constraints.
All the models $M1$--$M5$ are implemented with Python $3.9.23$ and PyTorch $2.0.0$.

To ensure a fair comparison, the rank of  $M1$ is set to 5, and the embedding dimension for each input feature in $M2$,$ M3$,$ M4$, and $M5$ is uniformly set to 5. Other hyperparameters of each model (such as the hidden layer dimensions in $M2$ and $M5$) are tuned using grid search to achieve optimal performance. To eliminate the effects of random parameter initializations, we independently ran each model 5 times with different random seeds (38, 40, 42, 44, 46) and calculated the mean and standard deviation of the performance metrics. The experimental results are summarized in  \ref{tab:results_summary}. Based on these results, this study draws the following important findings.

1) In overall performance (MAE, RMSE, $R^2$), $M1$ shows remarkable superiority (Table \ref{tab:results_summary}), achieving the lowest MAE/RMSE and highest $R^2$ among all compared models. This advantage comes from optimized input processing and a four-dimensional Tucker tensor—its Tucker-based high-order architecture captures complex turbulence dependencies, breaking linear/shallow model bottlenecks. In contrast, $M5$ has competitive MAE but significantly lower $R^2$ than $M1$; even with explicit interactions/deep features, it lacks explanatory power, confirming that tailored input adaptation and high-order tensor modeling are key to better performance.

2) $M1$'s five-fold cross-validation further verifies its effectiveness: in highly sparse three-dimensional wind field datasets, it demonstrates strong learning for known turbulence data and excellent prediction for unknown data. With targeted input processing and high-order tensor modeling, $M1$ acts as an efficient low-rank tensor completion model for turbulence analysis, outperforming common regression models and other compared models comprehensively.

\section{Conclusion}

In this paper, a NeuTucF model with a discretization module is proposed for turbulence regression. A discretization module is designed to partition the continuous three-dimensional wind data space into finite intervals, and a Tucker interaction layer is designed to model the complex interaction between the discretized wind field parameters and the Richardson number. The NeuTucF model achieves significant improvements in turbulence analysis in real three-dimensional wind field data than state-of-the-art models do. The discretization module is a generic tool and it can assist the NeuTucF or other similar models to process continuous input data. And the Tucker interaction layer is compatible with other common neural network architectures on turbulence analysis. In the future, we will study further the complex relationship between three-wind filed and turbulence indices with extending NeuTucF to improve accuracy of the turbulence prediction.

\bibliographystyle{unsrt} 
\bibliography{reference} 

\end{document}